\documentclass{article}
\usepackage{ijcai19}

\usepackage{times}
\usepackage{soul}
\usepackage{url}
\usepackage[hidelinks]{hyperref}
\usepackage[utf8]{inputenc}
\usepackage[small]{caption}
\usepackage{graphicx}
\usepackage{amsmath}
\usepackage{booktabs}
\usepackage{algorithm}
\usepackage{algorithmic}
\usepackage{textcomp}
\usepackage{enumitem}

\title{FAT-DeepFFM: Field Attentive Deep Field-aware Factorization Machine}

\author{
Junlin Zhang
\and
Tongwen Huang
\and
Zhiqi Zhang
\affiliations
Sina Weibo, Beijing, China
\emails
\{junlin6,tongwen,zhiqizhang\}@staff.weibo.com
}

\begin{document}

\maketitle

\begin{abstract}
Click through rate (CTR) estimation is a fundamental task in personalized advertising and recommender systems. Recent years have witnessed the success of both the deep learning based model and attention mechanism in various tasks in computer vision (CV) and natural language processing (NLP). How to combine the attention mechanism with deep CTR model is a promising direction because it may ensemble the advantages of both sides. Although some CTR model such as Attentional Factorization Machine (AFM) has been proposed to model the weight of second order interaction features, we posit the evaluation of feature importance before explicit feature interaction procedure is also important for CTR prediction tasks because the model can learn to selectively highlight the informative features and suppress less useful ones if the task has many input features. In this paper, we propose a new neural CTR model named Field Attentive Deep Field-aware Factorization Machine (FAT-DeepFFM) by combining the Deep Field-aware Factorization Machine (DeepFFM) with Compose-Excitation network (CENet) field attention mechanism which is proposed by us as an enhanced version of Squeeze-Excitation network (SENet) to highlight the feature importance. We conduct extensive experiments on two real-world datasets and the experiment results show that FAT-DeepFFM achieves the best performance and obtains different improvements over the state-of-the-art methods. We also compare two kinds of attention mechanisms (attention before explicit feature interaction vs. attention after explicit feature interaction) and demonstrate that the former one outperforms the latter one significantly.
\end{abstract}

\section{Introduction}
CTR estimation is a fundamental task in personalized advertising and recommender systems. Many models have been proposed to resolve this problem such as Logistic Regression (LR)\cite{mcmahan2013ad}, Polynomial-2 (Poly2) \cite{juan2016field}, tree-based models \cite{he2014practical}, tensor-based models\cite{koren2009matrix}, Bayesian models\cite{graepel2010web}, and Field-aware Factorization Machines (FFMs) \cite{juan2016field}. 
Deep learning techniques have shown promising results in many research fields such as computer vision\cite{krizhevsky2012imagenet,he2016deep}, speech recognition\cite{graves2013speech} and natural language understanding\cite{mikolov2010recurrent,cho2014learning}. As a result, employing DNNs for CTR estimation has also been a research trend in this field\cite{zhang2016deep,cheng2016wide,xiao2017attentional,guo2017deepfm,lian2018xdeepfm,wang2017deep,zhou2018deep,He2017NFM}. Some deep learning based models have been introduced and achieved success such as Factorisation-Machine Supported Neural Networks(FNN)\cite{zhang2016deep}, Attentional  Factorization Machine (AFM)\cite{xiao2017attentional},wide \& deep\cite{cheng2016wide},DeepFM\cite{guo2017deepfm} etc. On the other side, Attention mechanism can filter out the uninformative features from raw inputs and attention-based model has also been widely used and shown promising results on various tasks. How to combine the attention mechanism with deep CTR model is a promising direction because it may ensemble the advantages of both sides. Although some CTR model such as Attentional Factorization Machine\cite{xiao2017attentional} (AFM) has been proposed to model the weight of second order interaction features, we think the feature importance evaluation before explicit feature interaction is also important for CTR prediction tasks because the model can learn to selectively highlight the informative features and suppress less useful ones if the task has many input features.

In this work, we propose a new neural CTR model named Field ATtentive Deep Field-aware Factorization Machines (FAT-DeepFFM) by combining the neural field-aware factorization machines \cite{yang2017} with field attention mechanism. Specifically, what we do in this work is to introduce the Compose-Excitation network (CENet) which is an enhanced version of Squeeze-Excitation network (SENet) \cite{hu2017squeeze} like attention into DeepFFM model in order to improve the representational ability of deep CTR network. We aim to dynamically capture each feature's importance by explicitly modeling the interdependencies among all different features of an input instance before factorization machines's feature interaction procedure. Our goal is to use the CENet like attention mechanism to perform feature recalibration through which it can learn to selectively highlight the informative features and suppress less useful ones effectively.

The contributions of our work are summarized as follows:
\begin{enumerate}[label=\arabic*)]
\item 	We propose a novel model named FAT-DeepFFM that enhances the DeepFFM model by introducing the CENet  field attention to dynamically capture each feature's importance before explicit feature interaction procedure.
\item We compare two different kinds of attention mechanisms(attention on features before explicit feature interaction vs. attention on cross features after explicit feature interaction ) and the experiment results demonstrate that the former one outperforms the latter one significantly. 
\item We conduct extensive experiments on two real-world datasets and the experiment results show that FAT-DeepFFM achieves the best performance and obtains different improvements over the state-of-the-art methods.
\end{enumerate}

The rest of this paper is organized as follows. Section 2 introduces some related works which are relevant with our proposed model. We introduce our proposed Field Attentive Deep Field-aware Factorization Machine (FAT-DeepFFM) model in detail in Section 3. The experimental results on Criteo and Avazu datasets are presented and discussed in Section 4. Section 5 concludes our work in this paper. 

\section{Related Work}
\subsection{Factorization Machines and Field-aware Factorization Machine}
Factorization Machines (FMs) \cite{rendle2010factorization} and Field-aware Factorization Machines (FFMs) \cite{juan2016field} are two of the most successful CTR models. FMs use the dot product of two embedding vectors to model the effect of pairwise feature interactions. FFMs extended the ideas of Factorization Machines by additionally leveraging the field information and won two competitions hosted by Criteo and Avazu. When one feature interacts with other features from different fields, FFMs will learn different embedding vectors for each feature.

\subsection{Deep Learning based CTR Models}
With the great success of deep learning in many research fields such as Computer Vision and Natural language processing, many deep learning based CTR models have also been proposed in recent years. How to effectively model the feature interactions is the key factor for most of these neural network based models.

Factorisation-Machine Supported Neural Networks (FNN)\cite{zhang2016deep} is a feed-forward neural network using FM to pre-train the embedding layer. However, FNN can capture only high-order feature interactions. Wide \& Deep Learning\cite{cheng2016wide} was initially introduced for App recommendation in Google play. Wide \& Deep Learning jointly trains wide linear models and deep neural networks to combine the benefits of memorization and generalization for recommender systems. However, expertise feature engineering is still needed on the input to the wide part of Wide \& Deep model, which means that the cross-product transformation also requires to be manually designed. To alleviate manual efforts in feature engineering, DeepFM\cite{guo2017deepfm} replaces the wide part of Wide \& Deep model with FM and shares the feature embedding between the FM and deep component.  DeepFM is regarded as one state-of-the-art model in CTR estimation field.

Deep \& Cross Network (DCN)\cite{wang2017deep} efficiently captures feature interactions of bounded degrees in an explicit fashion. Similarly, eXtreme Deep Factorization Machine (xDeepFM) \cite{lian2018xdeepfm} also models the low-order and high-order feature interactions in an explicit way by proposing a novel Compressed Interaction Network (CIN) part.

Our approach is based on neural FFM which was firstly proposed by Yang\cite{yang2017} in Tencent Social Ads contest . It can be regarded as replacing the FM part of DeepFM with FFM and we will describe the model in detail in section 3.

\subsection{Attentive CTR Models}
Attention mechanism is motivated by human visual attention and it can filter out the uninformative features from raw inputs by reducing the side effects of noisy data. Attention-based model has been widely used and shown promising results on tasks such as speech recognition and machine translation. Attention mechanism is also introduced in some CTR models. For example, Attentional  Factorization Machine (AFM)\cite{xiao2017attentional} improves FM by discriminating and learning the importance of different feature interactions from data via a neural attention network. DIN\cite{zhou2018deep} represents users' diverse interests with an interest distribution and designs an attention-like network structure to locally activate the related interests according to the candidate ad.

\section{Field Attentive DeepFFM}
\subsection{DeepFFM}
Our work initially aims at introducing the FFM model into neural CTR systems. However, a similar effort to ours has been reported by Yang etc. \cite{yang2017} in Tencent Social Ads competition 2017. The authors report substantial gains after using neural FFM in their CTR prediction system. Neural FFM was quite successful in that competition: the 3rd place winner solution was based on this single Model and the ensemble version won the 1rd place in the competition. Because it's hard to find the detailed technical descriptions about this model, we will firstly introduce the neural FFM which will be called DeepFFM model in this paper. 

As we all know, FMs\cite{rendle2010factorization} model interactions between features i and j as the dot products of their corresponding embedding vectors as follows: 
\begin{equation}
     \hat{y}(x)= w_0 + \sum_{i=1}^{m}w_ix_i + \sum_{i=1}^{m}\sum_{j=i+1}^{m}\langle v_i,v_j\rangle x_ix_j
\end{equation}
An embedding vector $v_i \in R^k$ for each feature is learned by FM, k is a hyper-parameter which is usually a small integer and m is the feature number. However, FM neglects the fact that a feature might behave differently when it interacts with features from other fields. To explicitly take this difference into consideration, Field-aware Factorization Machines (FFMs) learn extra n-1 embedding vectors for each feature(here n denotes field number):
\begin{equation}
    \hat{y}(x)= w_0 + \sum_{i=1}^{m}w_ix_i + \sum_{i=1}^{m}\sum_{j=i+1}^{m}\langle v_{ij},v_{ji}\rangle x_ix_j
\end{equation}

\noindent where $v_{ij} \in R^{k}$ denotes the embedding vector of the j-th entry of feature i when feature i is interacting with fields j. k is the embedding size.

\begin{figure}[hbt!]
\includegraphics[width=8cm, height=6cm]{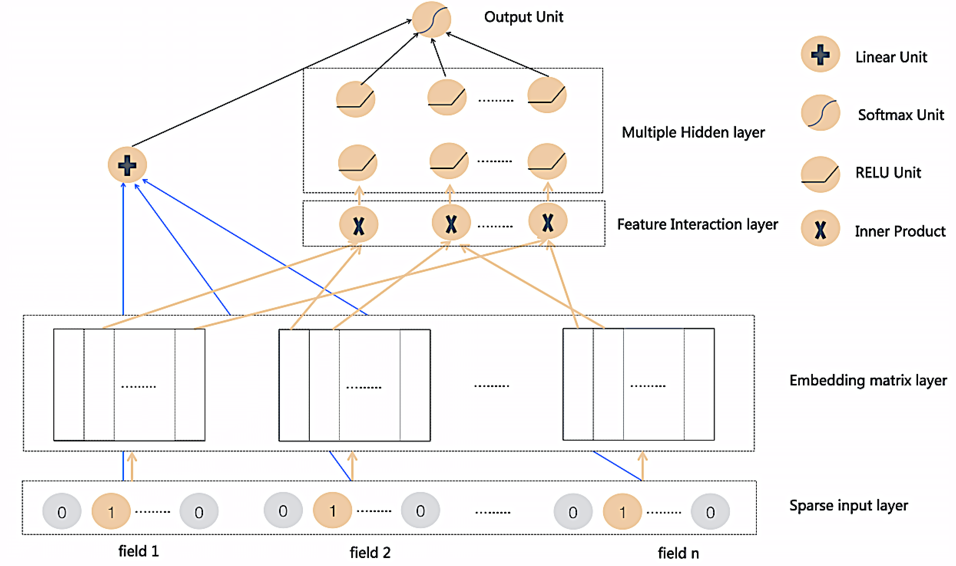}
\caption{The Neural Structure of Inner-Product version DeepFFM}
\label{fig:f1}
\end{figure}

As depicted in Figure 1, DeepFFM is designed to embody the idea of FFM through neural network. An input instance is firstly transformed into a high-dimensional sparse features via one-hot encoding to denote the raw feature input. The following embedding matrix layer is fully connected with sparse input layer to compress a raw feature to a low dimensional, dense real-value matrix. Specifically, for feature i, a corresponding 2-dimensional embedding matrix $EM_i=[v_{i1}, v_{i2}, \cdots, v_{ij}, \cdots, v_{in}]$   with size $k \times n$ is used to measure its impact of interactions with other features, where $v_{ij} \in R^{k}$ refers to the j-th embedding vector of field i, n is the number of fields and k is the size of embedding vector. So it's obvious that embedding matrix layer $EM$ is a 3-dimensional matrix with size $k \times n \times n$  because we have n fields and each field has one corresponding 2-dimensional embedding matrix.

The following feature interaction layer tries to capture the two way feature interactions between any pair of features from different fields on the embedding matrix $EM$. Denoting the feature interaction layer as vector $A$, we have two different types of feature interaction approaches: inner-product version and Hadamard-product version. We can formalize two methods in this layer as follows: \\
$A=[v_{12} \oplus v_{21}, \cdots, v_{ij} \oplus v_{ji}, \cdots,v_{(n-1)n} \oplus v_{n(n-1)}]   \  Inner \ Product$ \\
$A=[v_{12} \otimes v_{21}, \cdots, v_{ij} \times v_{ji}, \cdots, v_{(n-1)n} \otimes v_{n(n-1)}] \ Hadamard\ Product$\\
\noindent where $n$ is field number, $v_{ij} \oplus v_{ji}$ means the inner product of two embedding vectors as a scalar $\langle v_{ij},v_{ji} \rangle$ and $v_{ij} \times v_{ji}$ refers to the Hadamard product of two embedding vectors as following vector:

$v_{ij} \oplus v_{ji} = [v_{ij}^{1}\cdot v_{ji}^{1},v_{ij}^{2}\cdot v_{ji}^{2},\cdots,v_{ij}^{k}\cdot v_{ji}^{k}]$

\noindent where k is the size of embedding vector $v_{ji}$. Notice that $j>i$ is required in order to avoid the repeated computation.   We can see from here that feature interaction layer $A$ is a wide concatenated vector and the size of this vector is $n(n-1)/2$ if we adopt inner-product version while the size is $kn(n-1)/2$ if the Hadamard product version is adopted.

Multiple hidden layer is a feed-forward neural network on the feature interaction layer to implicitly learn high-order feature interactions. Denote the output of the feature interaction layer as vector  and we can feed it into hidden layer of feed-forward neural network. So the forward process is :
\begin{equation}
    x^1 = \sigma(W^1A+b^1)
\end{equation}
\begin{equation}
    x^l = \sigma(W^lx^{1-1}+b^l)
\end{equation}
where $l$ is the layer depth, $\sigma$ is an activation function, and $x^l$ is the output of the $l$-th hidden layer.

 Adding the linear part, the output unit of DeepFFM behaves as follows:
\begin{equation}
    \hat{y}(X) = \sigma(W_{1inear}x_{linear}+W^{l+1}x^{l}+b^{l+1})
\end{equation}
where $\sigma$ is the sigmoid function, $x_{linear}$ is the raw features, $x^l$ is the output of multiple hidden layer, $W_{1inear}$, $W^{l+1}$ and $b^{l+1}$ are learnable parameters.

\subsection{CENet Field Attention on Embedding Matrix Layer}
Hu proposed the ``Squeeze-and-Excitation Network'' (SENet) \cite{hu2017squeeze} to improve the representational power of a network by explicitly modeling the interdependencies between the channels of convolutional features in various image classification tasks. The SENet is proved to be successful in image classification tasks and won first place in ILSVRC 2017 classification task.

Our work is inspired by SENet's success in computer vision field. To improve the representational ability of deep CTR network, we introduce the Compose-Excitation network (CENet)  attention mechanism which is an enhanced version of SENet into DeepFFM model on embedding matrix Layer. We aim to dynamically capture each feature's importance by explicitly modeling the interdependencies among all different features before FM's feature interaction procedure. Our goal is to use the CENet attention mechanism to perform feature recalibration through which it can learn to selectively highlights the informative features and suppress less useful ones.

\begin{figure}[hbt!]
\includegraphics[width=8cm, height=6cm]{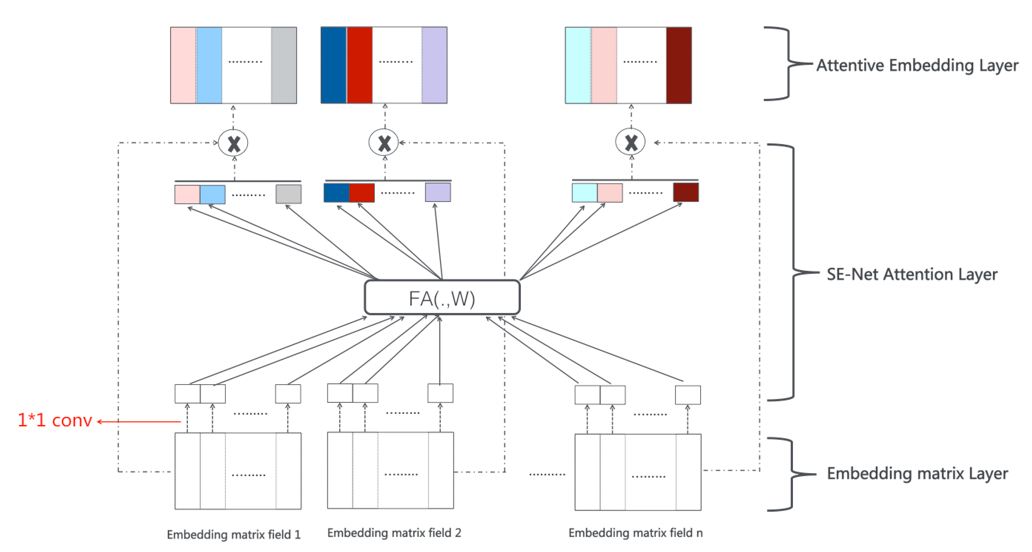}
\caption{CENet Field Attention}
\label{fig:f2}
\end{figure}

It can be seen from Figure 2 that the CENet like field attention mechanism involves two phases: Compose phase and Excitation phase. The first phase calculates ``summary statistics'' of each embedding vector of each field by composing all the information of one embedding vector into a simple feature descriptor; the second phase applies attentive transformations to these feature descriptors and then rescales the original embedding matrix using the calculated attention values. 

\noindent\textbf{Compose  Phase:}
Let $EM_i=[v_{i1},v_{i2},\cdots,v_{ij},\cdots,v_{in}]$ denotes the 2-dimensional $k \times n$ embedding matrix of field i, where $v_{ij} \in R^k$ refers to the j-th embedding vector of field i, n is the number of fields and k is the size of embedding vector. In this phase we compose the embedding vector $v_{ij}$ into one single number to represent the summary information of the feature.  This can be achieved by using $1 \times 1$ convolution\cite{szegedy2016rethinking,chollet2017xception} to generate feature-wise statistics instead of those squeeze operations such as global max pooling or sum operation commonly used in SENet. The $1 \times 1$ convolution, also called a pointwise convolution, is responsible for building new features through computing linear combinations of one input feature embedding. In SENet, we generate a statistic vector $z \in R^n$ for field i by shrinking each embedding vector, where the f-th element of $z_i$ is $z_{if} \in R$ which can be calculated by:


\begin{equation}
z_{if} = F_{sq}(v_{if})= \max\limits_{1 \leq t \leq k}v_{if}^{t}\hspace{0.5cm}Global\ Max\ Pooling
\end{equation}
Here, k means the embedding size of each embedding vector.

The most commonly used squeeze operation is global max pooling in CV field which can capture the strongest feature in corresponding channel. We change the method in this phase by using $1 \times 1$ convolution because we posit each position in feature embedding vector is informative in CTR task. So the $1 \times 1$ convolution can introduce parameters to learn the composing weight of each position in feature embedding. The $1 \times 1$ convolution is calculated as follows:
\begin{equation}
    z_{if}=conv1d(U_{if}, v_{if})=Relu(U_{if}v_{if})
\end{equation}
where $U_{if}$ is the convolution weight, the size of convolution kernal is $1 \times 1$, the number of filters is 1 and the activation function is set to 'Relu'.

\noindent\textbf{Excitation phase:}
After the first phase, the embedding matrix of field i $EM_i=[v_{i1},v_{i2},\cdots,v_{ij},\cdots,v_{in}]$ has been transformed into a descriptor vector $DV_i=[z_{i1},z_{i2},\cdots,z_{ij},\cdots,z_{in}]$ . We have n different fields, so we summarize all the descriptors by concatenating each descriptor vector as follows:
\begin{equation}
    D = concate(DV_1, DV_2, \cdots, DV_n)
\end{equation}            
 where the size of vector D is $n^2$.

 To calculate the attention from descriptor vector , two fully connected (FC) layers are used. The first FC layer is a dimensionality-reduction layer with parameters $W_1$ with reduction ratio $r$ which is a hyper-parameter and it uses ReLU as nonlinear function. The second FC layer increases dimensionality with parameters $W_2$ , which is equal to dimension of descriptor vector $D$ and it also uses ReLU as nonlinear activation function. Formally, the field attention is calculated as follows:
\begin{equation}
    S=F_{ex}(D,W)=\delta(W_2\delta(W_1D))
\end{equation}
where $\delta$ refers to the ReLU function, $W_1 \in R^{\frac{n^2}{r} \times n^2}$ and $W_1 \in R^{n^2 \times \frac{n^2}{r}}$,the size of attention vector $S$ is $n^2$ .

The activation of the ReLU function is used as the final field attention value without softmax normalization operation because we want to encourage multiple features to be important instead of just few of them. Then the values in original embedding matrix $EM_i$ of field i are rescaled by the accordingly calculated field attention vector $S_i$ as follows:
\begin{equation}
AEM_i=F_{scale}(S_i,EM_i)=[S_{i1} \cdot v_{i1},\cdots,S_{ij} \cdot v_{ij}, \cdots, S_{in} \cdot v_{in}]
\end{equation}
where $F_{scale}(S_i,EM_i)$ refers to vector-wise multiplication between embedding vector $v_{ij}$ and the scalar $S_{ij}$. The bigger attention value $S_{ij}$ implies that the model dynamically identifies an important feature and this attention value is used to boost the original embedding vector $v_{ij}$. On the contrary, small attention value $S_{ij}$ will suppress the uninformative features or even noise by decreasing the values in the corresponding embedding vector $v_{ij}$.

After the compose phase and excitation phase, we have a new 3-dimensional embedding matrix $AEM$ with size $k \times n \times n$, which is equal to the size of the original embedding matrix $EM$. We call the new embedding matrix attentive embedding layer in our paper.

\subsection{Combining the field attention and DeepFFM}
\begin{figure}[hbt!]
\includegraphics[width=8cm, height=6cm]{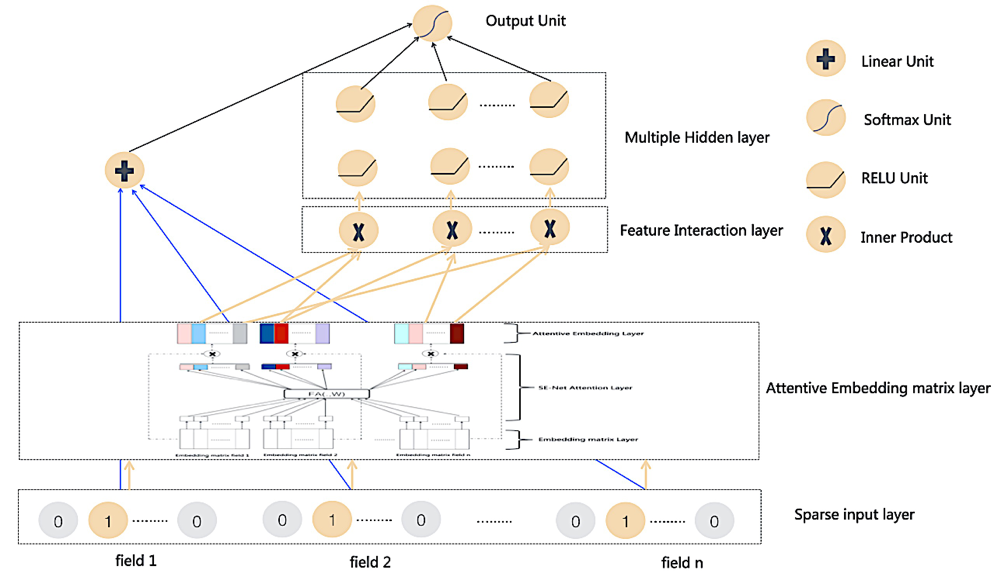}
\caption{Neural Structure of Field Attentive DeepFFM (Inner product version)}
\label{fig:f3}
\end{figure}
As discussed in Section 3.2, the CENet attention mechanism can perform feature recalibration through which it can learn to selectively highlights the informative features and suppress less useful ones. We can enhance DeepFFM model which is described in section 3.1 by inserting the CENet attention module into it. Figure 3 provides overall architecture of our proposed Field Attentive Deep Field-aware Factorization Machine (FAT-DeepFFM). It's similar in neural structure to DeepFFM while the original embedding matrix layer is replaced by the SE-Net like field attention module. We call this newly plugged-in module attentive embedding matrix layer. The other components of FAT-DeepFFM are same as the DeepFFM model. Similar to the DeepFFM, there are also two versions of FAT-DeepFFM according to the feature interaction type: inner-product version and Hadamard-product version.

We can see from the above-mentioned descriptions that our proposed attention mechanism is a kind of attention before cross features were produced. So a natural research question arises that which one will perform better if we introduce attention on cross features after the explicit feature interaction procedure just like AFM does? To answer this question, we also conduct some experiments to compare the performance difference of two kinds of attention mechanisms. The experimental results demonstrate that the attention before feature interaction outperforms the one after feature interaction consistently. We will discuss these experiments in detail in Section 4.3.

\section{Experimental Results}
To comprehensively evaluate our proposed method, we design some experiments to answer the following research questions:

\noindent\textbf{RQ1} Can our proposed FAT-DeepFFM outperform the state-of-the-art deep learning based CTR models? \\
\textbf{RQ2} Which attention mechanism (attention on features before explicit feature interaction vs. attention on cross features after explicit feature interaction) will perform better on the real world CTR datasets? \\
\textbf{RQ3} Which feature interaction method (Inner-Product vs. Hadamard-product) is more effective in neural network based CTR models?

\subsection{Experiment Setup}
\subsubsection{Datasets}
The following two data sets are used in our experiments:
\begin{enumerate}
    \item Criteo\footnote{http://labs.criteo.com/downloads/} Dataset. 
As a very famous public real world display ad dataset with each ad display information and corresponding user click feedback, Criteo data set is widely used in many CTR model evaluation.  There are 26 anonymous categorical fields and 13 continuous feature fields in Criteo data set. We split the data into training and test set randomly by 90\%:10\%.
   \item Avazu\footnote{http://www.kaggle.com/c/avazu-ctr-prediction} Dataset.
The Avazu dataset consists of several days of ad click-through data which is ordered chronologically. For each click data, there are 24 fields which indicate elements of a single ad impression. We split the data into training and test set randomly by 80\%:20\%.
\end{enumerate}

\begin{table}[]
\centering
\caption{Statistics of the evaluation datasets}
\label{table:t1}
\begin{tabular}{|l|l|l|l|}
\hline
Datasets & \#Instances & \#Fields & \#Features \\ \hline
Criteo   & 45M        & 39      & 2.3M       \\ \hline
Avazu    & 40.43M     & 24      & 0.64M     \\ \hline
\end{tabular}
\end{table}

Table 1 lists the statistics of the evaluation datasets. For these two datasets, a small improvement in prediction accuracy is regarded as practically significant because it will bring a large increase in a company's revenue if the company has a very large user base.

\subsubsection{Evaluation Metrics} 	AUC (Area Under ROC) and Logloss (cross entropy) are used in our experiments as the evaluation metrics. These two metrics are very popular for binary classification tasks. AUC is insensitive to the classification threshold and the positive ratio. AUC's upper bound is 1 and larger value indicates a better performance. Log loss measures the distance between two distributions and smaller log loss value means a better performance.

\subsubsection{Models for Comparisons}
We compare the performance of the following CTR estimation models as baseline:LR, FM, FFM, FNN, DeepFM, AFM, Deep\&Cross Network(DCN) , xDeepFM and DeepFFM, all of which are discussed in Section 2 and Section 3. 

\subsubsection{Implementation Details}
We implement all the models with Tensorflow in our experiments. For optimization method, we use the Adam with a mini-batch size of 1000 and a learning rate is set to 0.0001.  Focusing on neural networks structures in our paper, we make the dimension of field embedding for all models to be a fixed value of 10. For models with DNN part, the depth of hidden layers is set to 3, the number of neurons per layer is 1600 for FFM-related models and 400 for all other deep models, all activation function are ReLU and dropout rate is set to 0.5. For CENet component, the activation function is ReLU and the reduction ratio is set to 1 in all the related experiments. We conduct our experiments with 2 Tesla K40 GPUs.

\subsection{Performance Comparison (RQ1)}
\begin{table}[]
\centering
\caption{Overall performance of different models on Criteo and Avazu(the model name with suffix ``I'' means inner-product version while with suffix ``H'' means Hadamard product version)}
\label{table:t2}
\begin{tabular}{lllll}
\toprule
                       & \multicolumn{2}{c}{Criteo}        & \multicolumn{2}{c}{Avazu} 
                       \\\hline
Model Name             & AUC             & Logloss 
& AUC             & Logloss \\
\hline
LR                     & 0.7808          & 0.4681          & 0.7633          & 0.3891          \\
FM                     & 0.7923          & 0.4584          & 0.7745          & 0.3832          \\
FFM                    & 0.8001          & 0.4525          & 0.7795          & 0.3810           \\
FNN                    & 0.8057          & 0.4464          & 0.7802          & 0.3800            \\
AFM                    & 0.7965          & 0.4541          & 0.7740           & 0.3839          \\
DeepFM                 & 0.8085          & 0.4445          & 0.7786          & 0.3810           \\
DCN                    & 0.7977          & 0.4617          & 0.7680           & 0.3940           \\
xDeepFM                & 0.8091          & 0.4461          & 0.7808          & 0.3819          \\
\hline
DeepFFM-I & 0.8087 & 0.4434 & 0.7839 & 0.3783 \\
DeepFFM-H & 0.8088 & 0.4434 & 0.7835 & 0.3782 \\

\hline
\textbf{FAT-DeepFFM-I} & 0.8099          & 0.4421          & 0.7857          & 0.3763 \\
\textbf{FAT-DeepFFM-H} & \textbf{0.8104} & \textbf{0.4416} & \textbf{0.7863} & \textbf{0.3761}  \\\hline
\end{tabular}
\end{table}

The overall performance for CTR prediction of different models on Criteo dataset and Avazu dataset is shown in Table 2. We have the following key observations:
\begin{enumerate}
\item FAT-DeepFFM achieves the best performance in general and obtains different improvements over the state-of-the-art methods. As the best model, FAT-DeepFM outperforms FM by 3.64\% and 1.80\% in terms of Logloss (2.28\% and 1.50\% in terms of AUC) and outperforms LR by 5.64\% and 3.29\% in terms of Logloss (3.79\% and 2.99\% in terms of AUC) on Criteo and Avazu datasets. 
\item	FAT-DeepFFM consistently outperforms DeepFFM on both          datasets. This indicates that CENet field attentive mechanism is rather helpful for learning the importance of raw features.
\end{enumerate}

\subsection{Attention Mechanism Comparison (RQ2)}
In this subsection, we will discuss the performance of two different kinds of attention mechanisms: one is an attention before explicit feature interaction just like above-mentioned field attention; the other is the attention on cross features after explicit feature interaction procedure just like AFM does.

As for the specific approach for the attention on cross features, two methods are implemented: the similar CENet attention as described in section 3.2 and the MLP based attention just like AFM does, the hyper-parameters are tuned to achieve the best performance. The experimental results is shown in table 3 and the experiments with prefix ``MLP'' refer to the MLP based attention on cross features while the experiments with prefix ``CE'' mean the CENet attention is used on cross features.

Table 3 lists the overall performance of two attention mechanisms on Criteo dataset and Avazu dataset. We have the following key observations:
\begin{enumerate}
\item No matter which method (CENet or MLP based model as AFM does) is used as the specific attention approach on cross features, attention on features before feature interaction outperforms the attention on cross features after explicit feature interaction consistently, sometimes with a large margin. We infer it's perhaps because the attention on features highlights the important information while suppressed the unimportant features and noise more effectively compared with the attention on cross features. 
\item 	Under some conditions, the attention on cross features is harmful for some real world CTR prediction task. From the results of the inner-product group experiments shown in table 3, we can see that both the MLP-DeepFFM-I and CE-DeepFFM-I model underperform the original DeepFFM model. This result demonstrates that attention on cross features is harmful for CTR prediction task if we use inner-product function as the feature interaction method. The behind reason still needs further investigation.
\end{enumerate}

\subsection{Feature Interaction Method(RQ3)}


\begin{table}[]
\caption{Overall performance of two attention mechanisms on Criteo and Avazu(the model name with suffix ``I'' means inner-product version while with suffix ``H'' means Hadamard product version)}
\label{table:t3}
\begin{tabular}{lllll}
\hline
 & \multicolumn{2}{c}{Criteo} & \multicolumn{2}{c}{Avazu} \\ \hline
Model & AUC & Logloss & AUC & Logloss \\ \hline
DeepFFM-I & 0.8087 & 0.4434 & 0.7839 & 0.3783 \\
MLP-DeepFFM-I & 0.8022 & 0.4499 & 0.7819 & 0.3796 \\
CE-DeepFFM-I & 0.808 & 0.444 & 0.7816 & 0.381 \\
\textbf{FAT-DeepFFM-I} & \textbf{0.8099} & \textbf{0.4422} & \textbf{0.7857} & \textbf{0.3763} \\ \hline
DeepFFM-H & 0.8088 & 0.4434 & 0.7835 & 0.3782 \\
MLP-DeepFFM-H & 0.8083 & 0.444 & 0.7847 & 0.3778 \\
CE-DeepFFM-H & 0.8092 & 0.443 & 0.7822 & 0.3786 \\
\textbf{FAT-DeepFFM-H} & \textbf{0.8104} & \textbf{0.4417} & \textbf{0.7861} & \textbf{0.3773} \\ \hline
\end{tabular}
\end{table}

As we discussed in section 3, both the DeepFFM and FAT-DeepFFM model have two kinds of feature interaction approaches: inner product version vs. hadamard product version. So a natural research question arises that which approach will perform better? Table 3 also shows 4 groups of comparable experiments (model names with same prefix and different suffixes form one group such as DeepFFM-I and DeepFFM-H ) on two datasets. We have the following key observations:
\begin{enumerate}
    \item No apparent performance difference is observed if we don't adopt any attention to the DeepFFM model, no matter which feature interaction method is used (DeepFFM-I vs. DeepFFM-H).
    \item The Hadamard product function should be preferred to the inner product function if we adopt attention to the DeepFFM model, no matter attention on features or attention on cross features. We can see from table 3 that this conclusion holds in most cases.
\end{enumerate}

\section{Conclusion}
In this paper, we propose a new neural CTR model called Field Attentive Deep Field-aware Factorization Machine (FAT-DeepFFM) by combining the deep field-aware factorization machine (DeepFFM) with CENet field attention mechanism. We conduct extensive experiments on two real-world datasets and the experiment results show that FAT-DeepFFM achieves the best performance and obtains different improvements over the state-of-the-art methods. We also show that FAT-DeepFFM consistently outperforms DeepFFM on both datasets which indicates that CENet field attentive mechanism is rather helpful for learning the importance of raw features when the task has many input features. We also compare two different types of attention mechanisms (attention before explicit feature interaction vs. attention after explicit feature interaction) and the experiment results demonstrate that the former one outperforms the latter one significantly.

\bibliographystyle{named}
\bibliography{ijcai19}

\end{document}